# Celer: a Fast Solver for the Lasso with Dual Extrapolation

Mathurin Massias [1]   Alexandre Gramfort [1]   Joseph Salmon [2]

## Abstract

Convex sparsity-inducing regularizations are ubiquitous in high-dimensional machine learning, but solving the resulting optimization problems can be slow. To accelerate solvers, state-of-the-art approaches consist in reducing the size of the optimization problem at hand. In the context of regression, this can be achieved either by discarding irrelevant features (screening techniques) or by prioritizing features likely to be included in the support of the solution (working set techniques). Duality comes into play at several steps in these techniques. Here, we propose an extrapolation technique starting from a sequence of iterates in the dual that leads to the construction of improved dual points. This enables a tighter control of optimality as used in stopping criterion, as well as better screening performance of Gap Safe rules. Finally, we propose a working set strategy based on an aggressive use of Gap Safe screening rules. Thanks to our new dual point construction, we show significant computational speedups on multiple real-world problems.

## 1. Introduction

Following the seminal work on the Lasso (Tibshirani, 1996) (also known as Basis Pursuit (Chen & Donoho, 1995) in signal processing), convex sparsity-inducing regularizations have had a major impact on machine learning (Bach et al., 2012). Now thoroughly analyzed in terms of statistical efficiency (Bickel et al., 2009), these regularizers yield a sparse solution, meaning both a more interpretable model, as well as reduced time for prediction. In machine learning applications, the default method to optimize such problems is coordinate descent (Fu, 1998; Friedman et al., 2010).

Since by design only a few features are included in the optimal solution (what we will refer to as the *support*), state-of-the-art techniques rely on limiting the size of the (sub-)problems to consider. To do so, various approaches can be distinguished: *screening* techniques (Wang et al., 2013; Ogawa et al., 2013; Fercoq et al., 2015), following the seminal work of El Ghaoui et al. (2012), strong rules (Tibshirani et al., 2012) implemented in the GLMNET package or correlation screening (Xiang & Ramadge, 2012). Similar techniques discarding gradients have also been considered in the stochastic gradient descent literature (Vainsencher et al., 2015; Shibagaki et al., 2016). When a screening rule guarantees that all discarded features cannot be in the solution, it is called *safe*. The current state-of-the-art safe screening rules are the so-called Gap Safe rules (Ndiaye et al., 2017) relying on duality gap evaluation, which itself requires to know a suitable dual point.

Alternatively, *working sets*[1] (WS) techniques (Fan et al., 2008; Boisbunon et al., 2014; Johnson & Guestrin, 2015) select a subset of important features according to a particular criterion, and approximately solve the subproblem restricted to these features. A new subset is then defined, and the procedure is repeated. While screening techniques start from full problems and prune the feature set, WS techniques rather start with small problems and include more and more features if needed. For these techniques, duality can also come into play, both in the stopping criterion of the subproblem solver, as well as in the WS definition.

The organization of the paper is as follows: in Section 2, we remind the practical importance of duality for Lasso solvers and present a technique called dual extrapolation to obtain better dual points. We also shed some light on the success of our approach when combined with cyclic coordinate descent by interpreting the latter as Dykstra's algorithm in the Lasso dual. In Section 3, we show how dual extrapolation is well-suited to improve Gap Safe screening. We present in Section 4 a WS strategy based on an aggressive relaxation of the Gap Safe rules. Experiments in Section 6 show significant computational speedups on multiple real-world problems.

---

[1] INRIA, Université Paris-Saclay [2] LTCI, Télécom ParisTech, Université Paris-Saclay. Correspondence to: Mathurin Massias <mathurin.massias@inria.fr>.

---

[1] also called *active sets*; we choose *working* because in the screening literature, active set refers to the non discarded features.

**Notation**

For any integer $d$, $[d]$ denotes the set $\{1, \ldots, d\}$. We denote by $n$ the number of observations, by $p$ the number of features, $X = [x_1| \ldots |x_p] \in \mathbb{R}^{n \times p}$ represents the design matrix, and $y \in \mathbb{R}^n$ is the observation vector. For $\beta \in \mathbb{R}^p$, $\mathcal{S}_\beta := \{j \in [p] : \beta_j \neq 0\}$ is its support. For $\mathcal{W} \subset [p]$, $\beta_\mathcal{W}$ is $\beta$ restricted to the indices in $\mathcal{W}$, $X_\mathcal{W}$ is the matrix $X$ restricted to the columns with indices in $\mathcal{W}$. The vector $\mathbf{1}_K$ (resp. $\mathbf{0}_K$) has $K$ entries set to 1 (resp. 0). The norm $\|\cdot\|$ denotes the Euclidean norm on vectors or matrices. For $x \in \mathbb{R}$, $\text{sign}(x) = x/|x|$ (with the convention $\frac{0}{0} = 0$) and $\text{ST}(x, u) = \text{sign}(x) \cdot \max(0, |x| - u)$ is the soft-thresholding at level $u \in \mathbb{R}_+$. When applied to a vector, sign and ST act entry-wise. For any closed convex set $C$, we write $\Pi_C$ for the (Euclidean) projection onto $C$.

## 2. Duality for the Lasso

The Lasso estimator is defined as a[2] solution of

$$\hat{\beta} \in \underset{\beta \in \mathbb{R}^p}{\arg\min} \underbrace{\tfrac{1}{2} \|y - X\beta\|^2 + \lambda \|\beta\|_1}_{\mathcal{P}(\beta)} \ , \qquad (1)$$

where $\lambda > 0$ is a parameter controlling the trade-off between data-fitting and regularization.

The Lasso dual formulation reads, see Kim et al. (2007),

$$\hat{\theta} = \underset{\theta \in \Delta_X}{\arg\max} \underbrace{\tfrac{1}{2} \|y\|^2 - \tfrac{\lambda^2}{2} \left\|\theta - \tfrac{y}{\lambda}\right\|^2}_{\mathcal{D}(\theta)} \ , \qquad (2)$$

where $\Delta_X = \{\theta \in \mathbb{R}^n : \|X^\top \theta\|_\infty \leq 1\}$ is the (rescaled) dual feasible set. The associated duality gap is defined by $\mathcal{G}(\beta, \theta) := \mathcal{P}(\beta) - \mathcal{D}(\theta)$, for any primal-dual pair $(\beta, \theta) \in \mathbb{R}^p \times \Delta_X$. In particular, as illustrated in Figure 1a, the dual problem is equivalent to computing $\Pi_{\Delta_X}(y/\lambda)$.

**Proposition 1.** *Strong duality holds for Problem* (1)*, and primal and dual solutions verify:*

$$\hat{\theta} = \frac{1}{\lambda}(y - X\hat{\beta}) \ . \qquad (3)$$

*Moreover, $\mathcal{G}(\hat{\beta}, \hat{\theta}) = 0$.*

*Proof.* See for example Bauschke & Combettes (2011). □

### 2.1. Stopping iterative solvers

In general, Problem (1) does not admit a closed-form solution. Iterative optimization procedures such as (block) coordinate descent (BCD/CD) (Tseng, 2001; Friedman et al., 2007) (resp. ISTA/FISTA (Beck & Teboulle, 2009)) are among the most popular algorithms when dealing with

---

[2]recall that the solution might not be unique

---

**Algorithm 1** CYCLIC CD WITH DUAL EXTRAPOLATION
**input** : $X = [x_1| \ldots |x_p], y, \lambda, \beta^0, \epsilon$
**param**: $T, K = 5, f = 10$
**init** : $r = r^0 = y - X\beta^0, \theta^0 = r/\max(\lambda, \|X^\top r\|_\infty)$
**for** $t = 1, \ldots, T$ **do**
    **if** $t \equiv 0 \mod f$ **then** // $\theta$ every $f$ epoch only
        $s = t/f$         // dual point indexing
        $r^s = r$
        compute $\theta^s_{\text{res}}$ and $\theta^s_{\text{accel}}$ with eqs. (4) to (6)
        $\theta^s = \underset{\theta \in \{\theta^{s-1}, \theta^s_{\text{accel}}, \theta^s_{\text{res}}\}}{\arg\max} \mathcal{D}(\theta)$   // Eq (13)
        **if** $\mathcal{G}(\beta^t, \theta^s) < \epsilon$ **then**
            break
    **for** $j = 1, \ldots, p$ **do**
        $\beta_j^{t+1} = \text{ST}\left(\beta_j^t + \frac{x_j^\top r}{\|x_j\|^2}, \frac{\lambda}{\|x_j\|^2}\right)$
        **if** $\beta_j^{t+1} \neq \beta_j^t$ **then**
            $r \mathrel{+}= (\beta_j^t - \beta_j^{t+1})x_j$
**return** $\beta^t, \theta^s$

---

high dimensional applications in machine learning (resp. in image processing). A key practical question for iterative algorithms is the stopping criterion: when should the algorithm be stopped? Because strong duality holds for the Lasso, for any pair $(\beta, \theta) \in \mathbb{R}^p \times \Delta_X$, we have $\mathcal{P}(\beta) - \mathcal{P}(\hat{\beta}) \leq \mathcal{G}(\beta, \theta)$, which means that the duality gap provides an upper bound for the suboptimality gap. Therefore, given a tolerance $\epsilon > 0$, if at iteration $t$ of the algorithm we can construct $\theta^t \in \Delta_X$ s.t. $\mathcal{G}(\beta^t, \theta^t) \leq \epsilon$, then $\beta^t$ is guaranteed to be an $\epsilon$-optimal solution of (1).

Since Eq. (3) holds at optimality, a canonical choice of dual point relies on *residuals rescaling*. It consists in choosing a dual feasible point proportional to the residual $r^t := y - X\beta^t$, see for instance Mairal (2010):

$$\theta^t_{\text{res}} := r^t / \max(\lambda, \|X^\top r^t\|_\infty) \ . \qquad (4)$$

It is clear that if $\beta^t$ converges to $\hat{\beta}$, $\theta^t_{\text{res}}$ converges to $\hat{\theta}$, hence the duality gap for $(\beta^t, \theta^t_{\text{res}})$ goes to 0. Additionally, the cost of computing $\theta_{\text{res}}$ is moderate: $\mathcal{O}(np)$, the same as a single gradient descent step or an epoch of CD.

However, using $\theta_{\text{res}}$ has two noticeable drawbacks: it ignores information from previous iterates, and rescaling the residual $r^t$ makes an "unbalanced" use of computations in the sense that most of the burden is spent on improving $\beta$ while $\theta$ is obtained by solving a crude 1D optimization problem, *i.e.*, $\min \{\alpha \in [\lambda, +\infty] : r^t/\alpha \in \Delta_X\}$.

In practice (see Section 6), it turns out that, while safe and simple, such a construction massively overestimates the suboptimality gap, leading to slow safe feature identification and to numerical solvers running for more steps than actually needed. The new dual point construction we propose aims at improving upon this default strategy.

**Algorithm 2** DYKSTRA'S ALTERNATING PROJECTION

**input** : $\Pi_{C_1}, \ldots, \Pi_{C_p}, z$
**init** : $\theta = z, q_1 = 0, \ldots, q_p = 0$
**for** $t = 1, \ldots$ **do**
  **for** $j = 1, \ldots, p$ **do**
    $\tilde{\theta} \leftarrow \theta + q_j$
    $\theta \leftarrow \Pi_{C_j}(\tilde{\theta})$
    $q_j \leftarrow \tilde{\theta} - \theta$
**return** $\theta$

**Algorithm 3** DYKSTRA FOR THE LASSO DUAL

**input** : $X = [x_1|\ldots|x_p], y, \lambda$
**init** : $r = y, \tilde{\beta}_1 = 0, \ldots, \tilde{\beta}_p = 0$
**for** $t = 1, \ldots$ **do**
  **for** $j = 1, \ldots, p$ **do**
    $\tilde{r} \leftarrow r + x_j \tilde{\beta}_j$
    $r \leftarrow \tilde{r} - \text{ST}\left(\frac{x_j^\top \tilde{r}}{\|x_j\|^2}, \frac{1}{\|x_j\|^2}\right) \cdot x_j$
    $\tilde{\beta}_j \leftarrow \text{ST}\left(\frac{x_j^\top \tilde{r}}{\|x_j\|^2}, \frac{1}{\|x_j\|^2}\right)$
**return** $r/\lambda$

**Remark 1.** Other criteria than suboptimality are also often considered. For instance, the solver can be stopped as soon as the $\ell_2$ or $\ell_\infty$ norm of $\beta^t - \beta^{t-1}$ goes below a threshold $\epsilon$, or when the objective function stops decreasing fast ($\mathcal{P}(\beta^{t-1}) - \mathcal{P}(\beta^t) < \epsilon$). However, contrary to dual gap stopping rules, such heuristic rules do not offer a control on suboptimality. They are also tightly coupled with the value of the step size, making the use of a general $\epsilon$ difficult.

### 2.2. Dual extrapolation

Building on the work on nonlinear regularized acceleration by Scieur et al. (2016), we propose a new construction to obtain a better dual point. Instead of relying only on the last residual $r^t$, its approximation is improved by extrapolating previous residuals, *i.e.,* using $r^t, r^{t-1}, r^{t-2}$, etc.

**Definition 1** (Extrapolated dual point). For a fixed number of iterations $K$ (default being $K = 5$), let

$$r_{\text{accel}}^t = \begin{cases} r^t, & \text{if } t \leq K \\ \sum_{k=1}^{K} c_k r^{t+1-k}, & \text{if } t > K \end{cases} \quad (5)$$

where $c = (c_1, \ldots, c_K)^\top \in \mathbb{R}^K$ reads $c = z/(z^\top \mathbf{1}_K)$, and $z$ solves the linear system $(U^t)^\top U^t z = \mathbf{1}_K$ with $U^t = [r^{t+1-K} - r^{t-K}, \ldots, r^t - r^{t-1}] \in \mathbb{R}^{n \times K}$. Then,

$$\theta_{\text{accel}}^t := r_{\text{accel}}^t / \max(\lambda, \|X^\top r_{\text{accel}}^t\|_\infty). \quad (6)$$

This means that for the $K$ first iterations, $\theta_{\text{accel}}^t$ is equal to the classical dual point $\theta_{\text{res}}^t$. For subsequent iterations, the $K$ last values of the residuals are used to extrapolate the limit of the sequence $(r^t)$, and this extrapolation is rescaled to provide a feasible dual point.

**Remark 2.** The matrix $(U^t)^\top U^t$ may be singular, and solving the linear system may need Tikhonov regularization. This is discussed in Section 6.

**Theorem 1.** *When $r^t$ is obtained from iterations of ISTA, $\theta_{\text{accel}}^t$ converges to $\hat{\theta}$ as $t$ goes to $+\infty$.*

*Proof.* We recall that for $\mu = \|X\|_2^2$, the $t$-th iteration of ISTA is $\beta^{t+1} = \text{ST}(\beta^t + X^\top r^t/\mu, \lambda/\mu)$. The key result is that after some iterations, $\beta^t$ is a Vector AutoRegressive (VAR) process. This is true, since ISTA achieves *finite support identification* (Burke & Moré, 1988; Liang et al., 2014): if $\beta^t \to \hat{\beta}$, after a finite number of iterations, the sign is identified $\text{sign}(\beta^t) = \text{sign}(\hat{\beta})$ and so is the support $\mathcal{S}_{\beta^t} = \mathcal{S}_{\hat{\beta}} =: \hat{\mathcal{S}}$. In this regime, the non-linearity induced by the soft-thresholding becomes a simple bias:

$$\beta_{\hat{\mathcal{S}}}^{t+1} = \beta_{\hat{\mathcal{S}}}^t + \frac{1}{\mu} X_{\hat{\mathcal{S}}}^\top (y - X_{\hat{\mathcal{S}}} \beta_{\hat{\mathcal{S}}}^t) - \frac{\lambda}{\mu} \text{sign}(\beta_{\hat{\mathcal{S}}}^t)$$
$$= \beta_{\hat{\mathcal{S}}}^t + \frac{1}{\mu} X_{\hat{\mathcal{S}}}^\top (y - X_{\hat{\mathcal{S}}} \beta_{\hat{\mathcal{S}}}^t) - \frac{\lambda}{\mu} \text{sign}(\hat{\beta}_{\hat{\mathcal{S}}})$$
$$= \underbrace{(\text{Id}_p - \frac{1}{\mu} X_{\hat{\mathcal{S}}}^\top X_{\hat{\mathcal{S}}})}_{A} \beta_{\hat{\mathcal{S}}}^t + \underbrace{\frac{1}{\mu} X_{\hat{\mathcal{S}}}^\top y - \frac{\lambda}{\mu} \text{sign}(\hat{\beta}_{\hat{\mathcal{S}}})}_{b} .$$

Hence, $\beta_{\hat{\mathcal{S}}}^t$ is a (noiseless) VAR process: $\beta_{\hat{\mathcal{S}}}^{t+1} = A \beta_{\hat{\mathcal{S}}}^t + b$. It follows that $r^t = y - X_{\hat{\mathcal{S}}} \beta_{\hat{\mathcal{S}}}^t$ is also a VAR, and we can apply the result of Scieur et al. (2016, Prop. 2.2). □

Theorem 1 gives guarantees about $\theta_{\text{accel}}$ when $r^t$ is produced from ISTA iterates. In applications where $X$ is available as an explicit matrix and not as an implicit operator for which a fast transform exists (*e.g.,* FFT, wavelets), CD is more efficient (Friedman et al., 2007). While we do not have an equivalent of Theorem 1 for CD, on the datasets used in Section 6, we have observed excellent performance for dual extrapolation when combined with cyclic CD (see Section 6). Hence, we provide some insights on the efficiency of extrapolation for cyclic CD, described in Algorithm 1, by studying its connections with Dykstra's algorithm (Dykstra, 1983). To be precise, note that the points extrapolated are only the residual $r^t$ obtained every $f$ epochs: performing extrapolation at every CD update would be time consuming.

### 2.3. Dual perspective on CD

The Dykstra algorithm aims at solving problems of the form:

$$\hat{\theta} = \underset{\theta \in \cap_{j=1}^p C_j}{\arg \min} \|z - \theta\|^2 \quad (7)$$

where $C_1, \ldots, C_p$ are $p$ closed convex sets, with associated projections $\Pi_{C_1}, \ldots, \Pi_{C_p}$. The iterates of the (cyclic[3]) Dykstra algorithm are defined in Algorithm 2 (see Bauschke & Combettes (2011, Th. 29.2) for a convergence proof in the cyclic case).

The connection with CD for the Lasso has already been noticed (Tibshirani, 2017). In the Lasso dual, the closed convex sets are the $p$ slabs $C_j = \{\theta \in \mathbb{R}^n : -1 \leq x_j^\top \theta \leq 1\}$, and the point to be projected is $z = y/\lambda$. In this context, Dykstra's algorithm produces (non-necessarily feasible) iterates converging to $\hat{\theta}$.

The connection with CD can be made noticing that $(\mathrm{Id}_n - \Pi_{C_j})(\theta) = \mathrm{ST}(x_j^\top \theta / \|x_j\|^2, 1/\|x_j\|^2) x_j$. Using the change of variable $r = \lambda \theta$, $\tilde{r} = \lambda \tilde{\theta}$ and $q_j = x_j \beta_j / \lambda$ and the previous expression, Algorithm 2 is equivalent to Algorithm 3. It is to be noted that this is exactly cyclic CD for the Lasso, where the output $r$ of the algorithm corresponds to the residuals (and not the primal solution $\beta$).

On Figure 1, we illustrate Problem (2) for $n = p = 2$ (Figure 1a). Figure 1b (resp. Figure 1c) shows the iterates at the end of each epoch, produced by the cyclic (resp. shuffle) Dykstra algorithm, and their extrapolated version for $K = 4$. This corresponds to Algorithm 1 with $f = 1$. On Figure 1b, the iterates always lie on the same hyperplane, and they follow a noiseless VAR converging to $\hat{\theta}$: using only the last $K = 4$ points, extrapolation finds the true solution up to machine precision at the 5th iteration (Figure 1d). On the contrary, when the projection order on $C_1$ and $C_2$ is shuffled (Figure 1c), the iterates might not lie on the same hyperplane, and the trajectory tends to be less regular and harder to extrapolate. In what follows, we only consider cyclic orders due to their appealing interplay with extrapolation.

## 3. Gap Safe screening

The Lasso gives sparse solutions, meaning that $|\mathcal{S}_{\hat{\beta}}| \ll p$. Hence, if it were possible to discard features whose associated final coefficients vanish, the problem would become much smaller while having the same solutions. Discarding such features is called *screening*, and a key proposition for screening rules is the following:

$$\forall j \in [p], |x_j^\top \hat{\theta}| < 1 \Rightarrow \hat{\beta}_j = 0 \ . \quad (8)$$

Hence, the knowledge of $\hat{\theta}$ allows to identify the *equicorrelation set*[4] $\{j \in [p] : |x_j^\top \hat{\theta}| = 1\}$. The problem restricted to the equicorrelation set has the same solutions as (1), while being simpler : it typically has far less features. However, $\hat{\theta}$ is unknown so Equation (8) is not practical. To address this issue, Fercoq et al. (2015) have introduced the Gap Safe

---

[3] in the *shuffle* variant, the order is shuffled after each epoch
[4] even when primal solutions are not unique, the equicorrelation set contains the support of any solution (Tibshirani, 2013).

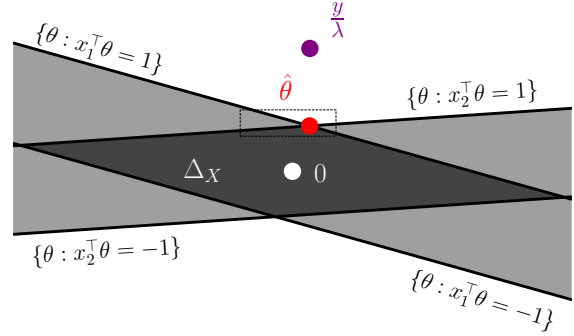

(a) Lasso dual problem with $X \in \mathbb{R}^{2 \times 2}$. A close-up on the dashed rectangle around $\hat{\theta}$ is given in Figure 1b and Figure 1c.

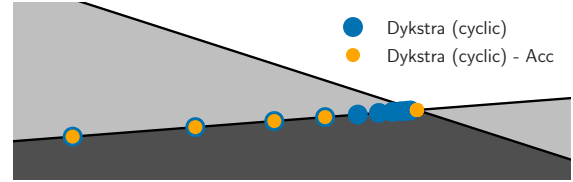

(b) Close up for cyclic Dykstra in the dual (end of each epoch)

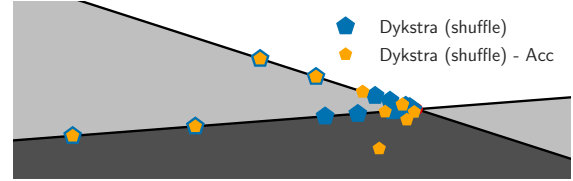

(c) Close up for shuffle Dykstra in the dual (end of each epoch)

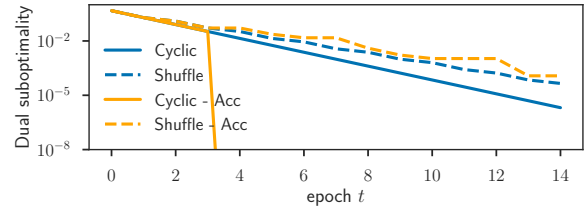

(d) Dual suboptimality with and without acceleration

Figure 1. In the Lasso case, the dual solution $\hat{\theta}$ is the projection of $y/\lambda$ onto the convex set $\Delta_X$ (the intersection of the two slabs).

rules to remove the $j$-th feature:

$$|x_j^\top \theta| < 1 - \|x_j\| \sqrt{\frac{2}{\lambda^2} \mathcal{G}(\beta, \theta)} \Rightarrow \hat{\beta}_j = 0, \quad (9)$$

which for any primal-dual feasible pair $(\beta, \theta) \in \mathbb{R}^p \times \Delta_X$, is *safe*, meaning that it will not wrongly discard a feature.

The Gap Safe rules have the appealing property of being convergent: at optimality, features not in the equicorrelation set have all been discarded. Additionally, they can be applied in a safe way in a sequential setting (Ndiaye et al., 2017) when only an approximate solution of Problem (1) is available for a $\lambda'$ close to $\lambda$ (e.g., for cross-validation).

Dynamic screening (Bonnefoy et al., 2014; 2015) is also possible using an iterate $\beta^t$ for a solver converging to $\hat{\beta}$: more and more features can be discarded along iterations.

Gap Safe rules performance depends strongly on how well $\theta$ approximates $\hat{\theta}$. Hence, $\theta$ acts as a certificate to discard irrelevant features: if the duality gap is large, the upper bound in (9) is crude, resulting in fewer (possibly) discarded features. Section 6.2 shows that $\theta_{\text{accel}}$ helps discarding more features than $\theta_{\text{res}}$, thus accelerating CD solvers and achieving safe feature identification in fewer epochs.

A potential drawback of screening rules is that, if the first duality gaps are large, CD computations are wasted on useless features during the first iterations (note that this is not the case when an approximation is available, e.g., when performing cross-validation; see Section 6.3). In the next section, we design a WS strategy to address this issue.

## 4. Working sets

Working set (WS) approaches involve two nested iteration loops: in the outer one, a set of features $\mathcal{W}_t \subset [p]$ is defined. In the inner one, an iterative algorithm is launched to solve the problem restricted to $X_{\mathcal{W}_t}$ (i.e., considering only the features in $\mathcal{W}_t$). In this section, we propose a WS construction based on an aggressive use of Gap Safe rules.

### 4.1. WS with aggressive gap screening

As it appears in Equation (9), the critical quantity measuring the importance of the $j$-th feature is

$$d_j(\theta) := \frac{1 - |x_j^\top \theta|}{\|x_j\|} \quad , \tag{10}$$

because

$$d_j(\theta) > \sqrt{\frac{2}{\lambda^2} \mathcal{G}(\beta, \theta)} \Rightarrow \hat{\beta}_j = 0 \quad . \tag{11}$$

Rather than discarding feature $j$ from the problem if $d_j(\theta)$ is too large, the WS is made of the coordinates achieving the lowest $d_j(\theta)$'s values. To do so, a first approach would consist in introducing a parameter $r \in ]0, 1[$ and creating a working set with features such that $d_j(\theta) < r\sqrt{2\mathcal{G}(\beta, \theta)/\lambda^2}$. However, a pitfall for this strategy is that the WS size is not explicitly under control: an inaccurate choice of $r$ could lead to extremely large WS, and would limit their benefits. Instead, to achieve a good control on the working set growth, we reorder the $d_j(\theta)$'s in a non-decreasing way: $d_{j_p}(\theta) \geq \cdots \geq d_{j_1}(\theta)$. Then, for a given working set size $p_t$, we choose, following Massias et al. (2017):

$$\mathcal{W}_t = \{j_1, \ldots, j_{p_t}\} \quad . \tag{12}$$

When $\theta = \theta_{\text{res}}^t$ and the features are normalized (a common, but not systematic preprocessing step), this WS construction simply consists in finding the $x_j$'s achieving the largest correlation with the residual, i.e., finding the largest $|x_j^\top r^t|$'s. Writing the data-fitting term $F(\beta) = \|y - X\beta\|^2/2$, and checking that $\nabla_j F(\beta^t) = -x_j^\top r^t$, then the previous rule coincides with gradient-based and correlation-based ones (Stich et al., 2017; Perekrestenko et al., 2017):

$$\begin{aligned} 1 - d_j(\theta_{\text{res}}^t) &= |x_j^\top r^t| / \max(\lambda, \|X^\top r^t\|_\infty) \\ &\propto |x_j^\top r^t| = |\nabla_j F(\beta^t)| \quad . \end{aligned}$$

However, the advantage of Equation (10) is that there is no restriction on the choice of the dual feasible point $\theta \in \Delta_X$. If a better candidate than rescaled residuals is available, it should be used instead. Considering the (ideal) case where the dual point constructed is $\hat{\theta}$, then the WS rule (12) yields the equicorrelation set (if $p_t$ is large enough), which is the best performance to expect in general for a WS construction.

When subproblems are solved with the same precision $\epsilon$ as considered for stopping the outer-loop and if the WS $\mathcal{W}_t$ grows geometrically (e.g., $p_{t+1} = 2p_t$) and monotonically (i.e., $\mathcal{W}_t \subset \mathcal{W}_{t+1}$), then convergence is guaranteed provided the inner solver converges. Indeed, this growth strategy guarantees that as long as the problem has not been solved up to precision $\epsilon$, more features are added, eventually starting the inner solver on the full problem until it reaches an $\epsilon$-solution. The initial WS size is set to $p_1 = 100$, except when an initialization $\beta^0 \neq \mathbf{0}_p$ is provided (e.g., for path/sequential computations, see Section 6.3), in which case we set $p_1 = |\mathcal{S}_{\beta^0}|$. This WS construction has many advantages: as it only requires a dual point, it is flexible and can be adapted to other objective functions (contrary to approaches such as Kim & Park (2010) which need to rewrite the Lasso as a QP). Moreover, exact resolution of the subproblems is not required for convergence. Our policy to choose $p_t$ avoids two common WS drawbacks: working sets growing one feature at a time, and cyclic behaviors, i.e., features entering and leaving the WS repeatedly.

We have coined our proposed algorithm implementing this WS strategy with dual extrapolation CELER (Constraint Elimination for the Lasso with Extrapolated Residuals).

## 5. Practical implementation

The implementation[5] is done in Python and Cython (Behnel et al., 2011).

**Linear system** If the linear system $(U^t)^\top U^t z = \mathbf{1}_K$ is ill-conditioned, rather than using Tikhonov regularization and solve $(U_t^\top U_t + \gamma I)z = \mathbf{1}_K$ as proposed in Scieur et al. (2016), we stop the computation for $\theta_{\text{accel}}$ and use $\theta_{\text{res}}$ for this iteration. In practice, this does not prevent the proposed

---
[5] https://github.com/mathurinm/celer

**Algorithm 4** CELER

**input** : $X, y, \lambda, \beta^0$
**param** : $p_{\text{init}} = 100, \epsilon, \underline{\epsilon} = 0.3, \text{max\_it}, \text{prune} = \text{True}$
**init** : $\theta^0 = \theta^0_{\text{inner}} = y / \|X^\top y\|_\infty$
**if** $\beta^0 \neq \mathbf{0}_p$ **then**   // warm start
$\quad p_1 = |\mathcal{S}_{\beta^0}|$
**else**
$\quad p_1 = p_{\text{init}}$
**for** $t = 1, \ldots, \text{max\_it}$ **do**
$\quad$ compute $\theta^t_{\text{res}}$
$\quad \theta^t = \arg\max_{\theta \in \{\theta^{t-1}, \theta^{t-1}_{\text{inner}}, \theta^t_{\text{res}}\}} \mathcal{D}(\theta)$   // Eq. (4)
$\quad g_t = \mathcal{G}(\beta^{t-1}, \theta^t)$   // global gap
$\quad$ **if** $g_t \leq \epsilon$ **then**
$\quad\quad$ break
$\quad$ **for** $j = 1, \ldots, p$ **do**
$\quad\quad$ compute $d^t_j = (1 - |x_j^\top \theta^t|)/\|x_j\|$
$\quad$ **if** prune **then**
$\quad\quad \epsilon_t = \underline{\epsilon} g_t$
$\quad\quad$ set $(d^t)_{\mathcal{S}_{\beta^{t-1}}} = -1$   // monotonicity
$\quad\quad$ **if** $t \geq 2$ **then**
$\quad\quad\quad p_t = \min(2|\mathcal{S}_{\beta^{t-1}}|, p)$   // Eq. (14)
$\quad$ **else**
$\quad\quad \epsilon_t = \epsilon$
$\quad\quad$ set $(d^t)_{\mathcal{W}_{t-1}} = -1$   // monotonicity
$\quad\quad$ **if** $t \geq 2$ **then**
$\quad\quad\quad p_t = \min(2p_{t-1}, p)$   // doubling size
$\quad \mathcal{W}_t = \{j \in [p] : d^t_j \text{ among } p_t \text{ smallest values of } d^t\}$
$\quad$ // Approximately solve sub-problem :
$\quad$ get $\tilde{\beta}^t, \theta^t_{\text{inner}}$ with Algorithm 1 applied to $(y, X_{\mathcal{W}_t}, \lambda, (\beta^{t-1})_{\mathcal{W}_t}, \epsilon_t)$
$\quad$ set $\beta^t = \mathbf{0}_p$ and $(\beta^t)_{\mathcal{W}_t} = \tilde{\beta}^t$
$\quad \theta^t_{\text{inner}} = \theta^t_{\text{inner}} / \max(\lambda, \|X^\top \theta^t_{\text{inner}}\|_\infty)$
**return** $\beta^t, \theta^t$

methodology from computing significantly lower gaps than the standard approach.

**Practical cost of dual extrapolation** The storage cost of dual extrapolation is $\mathcal{O}(nK)$ (storing $r^t, \ldots, r^{t-K}$). The main computation cost lies in the dual rescaling of $r^{\text{accel}}$, which is $\mathcal{O}(np)$, and corresponds to the same cost as an epoch of CD/ISTA. The cost of computing $c$ is small, since the matrix $(U^t)^\top U^t$ is only $K \times K$. One should notice that there is no additional cost to compute the residuals: in reasonable CD implementations, they have to be maintained at all iterations to avoid costly partial gradients computation (see Algorithm 1); for ISTA their computation is also required at each epoch to evaluate the gradients $X^\top r^t$. As usual for iterative algorithms, we do not compute the duality gap (nor the dual points) at every update of $\beta$, but rather after every $f = 10$ CD/ISTA epochs[6]. This makes the cost of dual extrapolation small compared to the iterations in the primal. The influence of $f$ and $K$ in practice is illustrated in additional experiments in Appendix A.1.

---

[6]This explains why the indices for $\beta$ and $\theta$ differ in Algorithm 1

**Robustifying dual extrapolation** Even if in practice we have observed fast convergence of $\theta^t_{\text{accel}}$ towards $\hat{\theta}$, Theorem 1 does not provide guarantees about the behavior of $\theta^t_{\text{accel}}$ when the residuals are constructed from iterates of CD or other algorithms. Hence, for a cost of $\mathcal{O}(np)$, in Algorithm 1 we also compute $\theta^t_{\text{res}}$ and use as dual point

$$\theta^t = \arg\max_{\theta \in \{\theta^{s-1}, \theta^s_{\text{accel}}, \theta^s_{\text{res}}\}} \mathcal{D}(\theta) \ . \quad (13)$$

The total computation cost of the dual is only doubled, which remains small compared to the cost of $f$ epochs of ISTA/CD, while guaranteeing monotonicity of the dual objective, and a behavior at least as good as $\theta^t_{\text{res}}$.

**Pruning** While the monotonic geometric growth detailed in Section 4 guarantees convergence, if $p_1$ is chosen too large, the working sets will never decrease. To remediate this, we introduce a variant called *pruning*:

$$p_t = \min(2|\mathcal{S}_{\beta^{t-1}}|, p) \ , \quad (14)$$

in which $\mathcal{W}_t$ approximately doubles its size at each iteration. This still guarantees that, even if $p_1$ was set too small, $p_t$ will grow quickly to reach the correct value. On the other hand, if $p_1$ is too big, many useless features are included at the first iteration, but is is likely that their coefficients will be 0, and hence $|\mathcal{S}_{\beta^1}|$ will be small, making $p_2$ small. This is illustrated by an experiment in Appendix A.2.

## 6. Experiments

### 6.1. Higher dual objective

We start by investigating the efficiency of our dual point in a case where $\lambda$ is fixed. Figure 2 shows, for the CD solver given in Algorithm 1, the duality gaps evaluated with the standard approach $\mathcal{P}(\beta^t) - \mathcal{D}(\theta^t_{\text{res}})$ and our proposed dual extrapolation $\mathcal{P}(\beta^t) - \mathcal{D}(\theta^t_{\text{accel}})$, as well as the exact suboptimality $\mathcal{P}(\beta^t) - \mathcal{P}(\hat{\beta})$ (note that the latter is not available to the practitioner before convergence). The experiment is performed on the *leukemia* dataset ($n = 72, p = 7,129$), with the design matrix columns set to unit $\ell_2$-norm, and $y$ centered and set to unit $\ell_2$-norm so that the first primal objective is $\mathcal{P}(\mathbf{0}_p) = 0.5$. The algorithm is run without warm start ($\beta^0 = \mathbf{0}_p$) for $\lambda = \lambda_{\max}/20$, and the values of $\theta^t_{\text{accel}}$ and $\theta^t_{\text{res}}$ are monitored[7]. $\mathcal{P}(\hat{\beta})$ is obtained by running the solver up to machine precision.

For a better analysis of the impact of dual extrapolation, in this experiment (and here only), we have not imposed

---

[7]$\lambda_{\max} := \|X^\top y\|_\infty$ is the smallest $\lambda$ s.t. $\hat{\beta} = \mathbf{0}_p$

monotonicity of the various dual objectives, nor have we used the best of both points as proposed in Equation (13).

As claimed in Section 2, we can observe that $\theta_{\text{res}}$ massively overestimates the suboptimality gap: while a true suboptimality gap of $10^{-6}$ is reached around epoch 200, the classical upper bound achieves this value at epoch 400 only. This means that if the duality gap were used as stopping criterion, the solver would run for twice too long. On the contrary, after a number of iterations where it behaves like the canonical approach, the proposed choice $\theta_{\text{accel}}$ accelerates and provides a duality gap much closer to the true suboptimality. After a sufficient number of epochs, the two are even almost equal, meaning that $\theta_{\text{accel}}^t$ is extremely close to $\hat{\theta}$. The difference between the two approaches is particularly striking for low values of $\epsilon$. We also see, that, although more bumpy than the standard approach, our proposed duality gap does not behave erratically. Hence, stabilizing it as stated in Equation (13) does not seem mandatory (but since it is cheap, we still do it for other experiments). Practical choices of $f$ and $K$ are discussed in Appendix A.1.

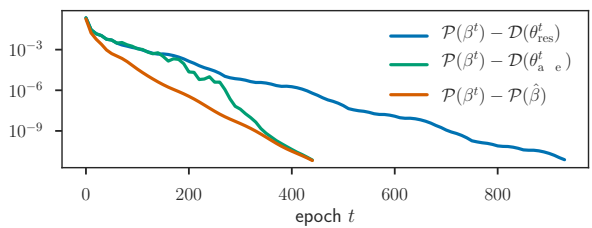

*Figure 2.* Duality gaps evaluated with the canonical dual point $\theta_{\text{res}}$ and the proposed construction $\theta_{\text{accel}}$, along with the true suboptimality gap. Performance is measured for Algorithm 1 on the *leukemia* dataset, for $\lambda = \lambda_{\max}/20$. Our duality gap quickly gets close to the true suboptimality, while the canonical approach constantly overestimates it.

### 6.2. Better Gap Safe screening performance

Figure 2 shows that extrapolated residuals yield tighter estimates of the suboptimality gap than rescaled residuals. However, one may argue either that using the duality gap as stopping criterion is infrequent (let us nevertheless mention that this criterion is for example the one implemented in the popular package `scikit-learn` (Pedregosa et al., 2011)), or that vanilla CD is seldom implemented alone, but rather combined with screening or working set techniques. Here we demonstrate the benefit of the proposed extrapolation when combined with screening: the number of screened features grows more quickly when our new dual construction is used. This leads to faster CD solvers, and quicker safe feature identification.

The dataset for this experiment is the Finance/E2006-log1p dataset (publicly available from LIBSVM[8]), preprocessed as follows: features with strictly less than 3 non-zero entries are removed, features are set to unit $\ell_2$-norm, $y$ is centred and set to unit $\ell_2$-norm, and an unregularized intercept feature is added. After preprocessing, $n = 16,087$ and $p = 1,668,738$.

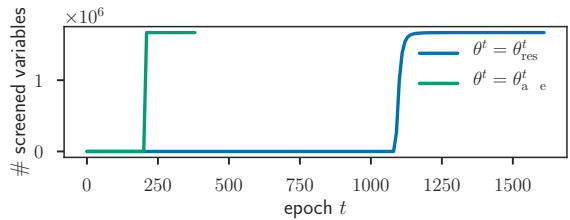

*Figure 3.* Number of variables discarded by the (dynamic) Gap Safe rule as a function of epochs of Algorithm 1, depending on the dual point used, for $\lambda = \lambda_{\max}/5$ (Finance dataset).

Figure 3 shows the number of screened variables as a function of the number of epochs in Algorithm 1, when using either standard residual rescaling or dual extrapolation to get the dual point $\theta^t$ in Equation (9). The solver stops once a duality gap of $10^{-6}$ is reached. We can see that the faster convergence of $\theta_{\text{accel}}^t$ towards $\hat{\theta}$ observed in Figure 2 translates into a better Gap Safe screening: features are discarded in fewer epochs than when $\theta_{\text{res}}^t$ is used. The gain in number of screened variables is directly reflected in terms of computation time: 70 s for the proposed approach, compared to 290 s for Gap Safe rule with rescaled residuals.

### 6.3. Working sets application to Lasso path

In practice, it rarely happens that the solution of Problem (1) must be computed for a single $\lambda$: the ideal value of the regularization parameter is not known, and $\hat{\beta}$ is computed for several $\lambda$'s, before the best is selected (*e.g.,* by cross-validation). The values of $\lambda$ are commonly[9] chosen on a logarithmic grid of 100 values between $\lambda_{\max}$ and $\lambda_{\max}/10^2$ or $\lambda_{\max}/10^3$. For the Finance dataset, we considered $\lambda_{\max}/10^2$, leading to a support of size 15,000. In such sequential context, warm start is standard and we implement it for all algorithms. It means that all solvers computing $\hat{\beta}$ are initialized with the approximate solution obtained for the previous $\lambda$ on the grid (starting from $\lambda_{\max}$).

A comparison between CELER and the BLITZ algorithm is presented in Figure 4. We refer to Johnson & Guestrin (2015) for a more extensive comparison which shows that BLITZ outperforms Lasso solvers such as L1_LS (Kim et al., 2007), APPROX (Fercoq & Richtárik, 2015) or GLMNET (Friedman et al., 2010) on a large collections of datasets and settings. In our experiments, we have used BLITZ's C++ open source implementation[10]. Results show that CELER clearly outperforms BLITZ. It also shows that the dynamic pruning of the working set does not bring much in this setting. Figure 10 in Appendix A.3 shows the same result for a coarser grid of 10 values of $\lambda$.

---

[8] http://www.csie.ntu.edu.tw/~cjlin/libsvmtools/datasets/.

[9] this is the default grid in GLMNET or scikit-learn

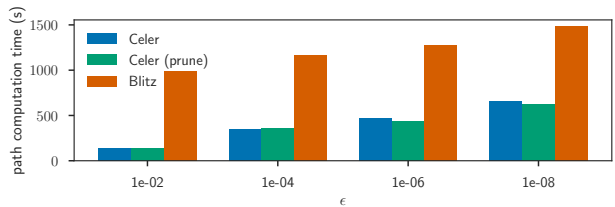

*Figure 4.* Times to solve the Lasso path to precision $\epsilon$ for 100 values of $\lambda$, from $\lambda_{\max}$ to $\lambda_{\max}/100$, on the Finance data. CELER outperforms BLITZ. Both safe and prune versions behave similarly.

**GLMNET comparison** Another popular solver for the Lasso is GLMNET, which uses working sets heuristics based on KKT conditions. However, the resulting solutions are not safe in terms of feature identification. Figure 5 shows that for the same value of stopping criterion[11], the supports identified by GLMNET contain much more features outside of the equicorrelation set (determined with Gap Safe rules after running CELER with $\epsilon = 10^{-14}$).

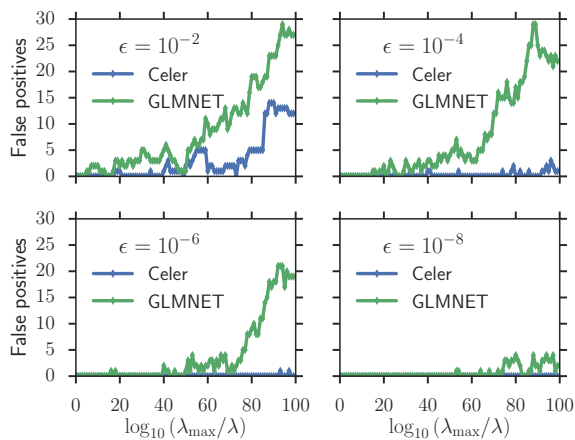

*Figure 5.* Number of false positives for GLMNET and CELER on a Lasso path on the *leukemia* dataset, depending on the stopping criterion $\epsilon$.

**Single $\lambda$** To demonstrate that the performance observed in Figure 4 is not only due to the sequential setting, we also perform an experiment for a single value of $\lambda = \lambda_{\max}/20$. The Lasso estimator is computed up to a desired precision $\epsilon$ which is varied between $10^{-2}$ and $10^{-6}$ (all solvers use duality gap). CELER is orders of magnitude faster than scikit-learn, which uses vanilla CD. The WS approach of BLITZ is also outperformed, especially for low $\epsilon$ values. This is most likely due to a bad estimation of the suboptimality gap with the dual point chosen in BLITZ.

---

[10] https://github.com/tbjohns/BlitzL1/
[11] on primal decrease for GLMNET, on duality gap for CELER

*Table 1.* Computation time (in seconds) for CELER, BLITZ and scikit-learn to reach a given precision $\epsilon$ on the Finance dataset with $\lambda = \lambda_{\max}/20$ (without warm start: $\beta^0 = \mathbf{0}_p$).

| $\epsilon$ | $10^{-2}$ | $10^{-3}$ | $10^{-4}$ | $10^{-6}$ |
|---|---|---|---|---|
| CELER | 5 | 7 | 8 | 10 |
| BLITZ | 25 | 26 | 27 | 30 |
| scikit-learn | 470 | 1350 | 2390 | - |

## 7. Discussion

The WS approach of BLITZ, analogous to that of CELER, is based on a geometric interpretation of the dual. The criterion to build $\mathcal{W}_t$ can be reformulated to match (12) (with the notable difference that $p_t$ is determined at runtime by solving an auxiliary optimization problem). However, for the analysis to hold, the dual point $\theta^t$ used in the outer loop must be a barycenter of the previous dual point $\theta^{t-1}$ and the current residuals, rescaled on the subproblem $r^{t-1}/\max(\lambda, \|X_{\mathcal{W}_{t-1}}^\top r^{t-1}\|_\infty)$. This prevents BLITZ from using extrapolation. The flexibility of CELER *w.r.t.* the choice of dual point enables it to benefit from the extrapolated dual point returned by the inner solver. Figure 4 shows that this dual point is key to outperform BLITZ.

For the sake of clarity and readability, we have focused on the Lasso case. Yet, the same methodology could be applied to generic $\ell_1$-type problems: $\hat{B} \in \arg\min_{B \in \mathbb{R}^{p \times q}} F(B) + \lambda \Omega(B)$, where $F$ is a smooth function and $\Omega$ is a norm separable over the rows of B. Many problems can be cast as such an instance: Multitask Lasso, Multiclass Sparse Logistic Regression, SVM dual, etc.

## Conclusion

In this paper we have illustrated the importance of improving duality gap computations for practical Lasso solvers. Using an extrapolation technique to create more accurate dual candidates, we have been able to accelerate standard solvers relying on screening and working set techniques. Our experiments on popular (sparse or dense) datasets showed the importance of dedicating some effort to the improvement of dual solutions: the combined benefits obtained both from improved stopping time and from screening accuracy has led to improved state-of-the-art solvers at little coding effort.

## Acknowledgments

This work was funded by ERC Starting Grant SLAB ERC-YStG-676943 and by the chair Machine Learning for Big Data of Télécom ParisTech.

## References

Bach, F., Jenatton, R., Mairal, J., and Obozinski, G. Convex optimization with sparsity-inducing norms. *Foundations and Trends in Machine Learning*, 4(1):1–106, 2012.

Bauschke, H. H. and Combettes, P. L. *Convex analysis and monotone operator theory in Hilbert spaces*. Springer, New York, 2011.

Beck, A. and Teboulle, M. A fast iterative shrinkage-thresholding algorithm for linear inverse problems. *SIAM J. Imaging Sci.*, 2(1):183–202, 2009.

Behnel, S., Bradshaw, R., Citro, C., Dalcin, L., Seljebotn, D. S., and Smith, K. Cython: The best of both worlds. *Computing in Science Engineering*, 13(2):31–39, 2011.

Bickel, P. J., Ritov, Y., and Tsybakov, A. B. Simultaneous analysis of Lasso and Dantzig selector. *Ann. Statist.*, 37(4):1705–1732, 2009.

Boisbunon, A., Flamary, R., and Rakotomamonjy. Active set strategy for high-dimensional non-convex sparse optimization problems. In *ICASSP*, pp. 1517–1521, 2014.

Bonnefoy, A., Emiya, V., Ralaivola, L., and Gribonval, R. A dynamic screening principle for the lasso. In *EUSIPCO*, 2014.

Bonnefoy, A., Emiya, V., Ralaivola, L., and Gribonval, R. Dynamic screening: accelerating first-order algorithms for the Lasso and Group-Lasso. *IEEE Trans. Signal Process.*, 63(19):20, 2015.

Burke, J. V. and Moré, J. J. On the identification of active constraints. *SIAM J. Numer. Anal.*, 25(5):1197–1211, 1988.

Chen, S. S. and Donoho, D. L. Atomic decomposition by basis pursuit. In *SPIE*, 1995.

Dykstra, R. L. An algorithm for restricted least squares regression. *J. Amer. Statist. Assoc.*, 78(384):837–842, 1983.

El Ghaoui, L., Viallon, V., and Rabbani, T. Safe feature elimination in sparse supervised learning. *J. Pacific Optim.*, 8(4):667–698, 2012.

Fan, R.-E., Chang, K.-W., Hsieh, C.-J., Wang, X.-R., and Lin, C.-J. Liblinear: A library for large linear classification. *J. Mach. Learn. Res.*, 9:1871–1874, 2008.

Fercoq, O. and Richtárik, P. Accelerated, parallel and proximal coordinate descent. *SIAM J. Optim.*, 25(3):1997–2013, 2015.

Fercoq, O., Gramfort, A., and Salmon, J. Mind the duality gap: safer rules for the lasso. In *ICML*, pp. 333–342, 2015.

Friedman, J., Hastie, T. J., Höfling, H., and Tibshirani, R. Pathwise coordinate optimization. *Ann. Appl. Stat.*, 1(2):302–332, 2007.

Friedman, J., Hastie, T. J., and Tibshirani, R. Regularization paths for generalized linear models via coordinate descent. *J. Stat. Softw.*, 33(1):1, 2010.

Fu, W. J. Penalized regressions: the bridge versus the lasso. *J. Comput. Graph. Statist.*, 7(3):397–416, 1998.

Johnson, T. B. and Guestrin, C. Blitz: A principled meta-algorithm for scaling sparse optimization. In *ICML*, pp. 1171–1179, 2015.

Kim, S.-J., Koh, K., Lustig, M., Boyd, S., and Gorinevsky, D. An interior-point method for large-scale $\ell_1$-regularized least squares. *IEEE J. Sel. Topics Signal Process.*, 1(4):606–617, 2007.

Kim, J. and Park, H. Fast active-set-type algorithms for l1-regularized linear regression. In *AISTATS*, pp. 397–404, 2010.

Liang, J., Fadili, J., and Peyré, G. Local linear convergence of forward–backward under partial smoothness. In *NIPS*, pp. 1970–1978, 2014.

Mairal, J. *Sparse coding for machine learning, image processing and computer vision*. PhD thesis, École normale supérieure de Cachan, 2010.

Massias, M., Gramfort, A., and Salmon, J. From safe screening rules to working sets for faster lasso-type solvers. In *NIPS-OPT*, 2017.

Ndiaye, E., Fercoq, O., Gramfort, A., and Salmon, J. Gap safe screening rules for sparsity enforcing penalties. *J. Mach. Learn. Res.*, 18(128):1–33, 2017.

Ogawa, K., Suzuki, Y., and Takeuchi, I. Safe screening of non-support vectors in pathwise SVM computation. In *ICML*, pp. 1382–1390, 2013.

Pedregosa, F., Varoquaux, G., Gramfort, A., Michel, V., Thirion, B., Grisel, O., Blondel, M., Prettenhofer, P., Weiss, R., Dubourg, V., Vanderplas, J., Passos, A., Cournapeau, D., Brucher, M., Perrot, M., and Duchesnay, E. Scikit-learn: Machine learning in Python. *J. Mach. Learn. Res.*, 12:2825–2830, 2011.

Perekrestenko, D., Cevher, V., and Jaggi, M. Faster coordinate descent via adaptive importance sampling. In *AISTATS*, pp. 869–877, 2017.

Scieur, D., d'Aspremont, A., and Bach, F. Regularized nonlinear acceleration. In *NIPS*, pp. 712–720, 2016.

Shibagaki, A., Karasuyama, M., Hatano, K., and Takeuchi, I. Simultaneous safe screening of features and samples in doubly sparse modeling. In *ICML*, pp. 1577–1586, 2016.

Stich, S., Raj, A., and Jaggi, M. Safe adaptive importance sampling. In *NIPS*, pp. 4384–4394, 2017.

Tibshirani, R. Regression shrinkage and selection via the lasso. *J. R. Stat. Soc. Ser. B Stat. Methodol.*, 58(1):267–288, 1996.

Tibshirani, R., Bien, J., Friedman, J., Hastie, T. J., Simon, N., Taylor, J., and Tibshirani, R. J. Strong rules for discarding predictors in lasso-type problems. *J. R. Stat. Soc. Ser. B Stat. Methodol.*, 74(2):245–266, 2012.

Tibshirani, R. J. The lasso problem and uniqueness. *Electron. J. Stat.*, 7:1456–1490, 2013.

Tibshirani, R. J. Dykstra's Algorithm, ADMM, and Coordinate Descent: Connections, Insights, and Extensions. In *NIPS*, pp. 517–528, 2017.

Tseng, P. Convergence of a block coordinate descent method for nondifferentiable minimization. *J. Optim. Theory Appl.*, 109(3):475–494, 2001.

Vainsencher, D., Liu, H., and Zhang, T. Local smoothness in variance reduced optimization. In *NIPS*, pp. 2179–2187, 2015.

Wang, J., Zhou, J., Wonka, P., and Ye, J. Lasso screening rules via dual polytope projection. In *NIPS*, pp. 1070–1078, 2013.

Xiang, Z. J. and Ramadge, P. J. Fast lasso screening tests based on correlations. In *ICASSP*, pp. 2137–2140, 2012.

## A. Additional experiments

### A.1. Choice of $f$ and $K$

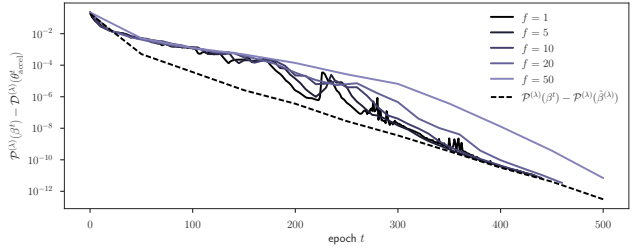

*Figure 6.* Duality gap evaluated with $\theta_{\text{accel}}$ as a function of the parameter $f$, for $K = 5$.

Figure 6 shows that if the residuals used to extrapolate are too close (small $f$), the performance of acceleration is too noisy (though the duality gap still converges to the true suboptimality gap). For residuals too far apart (large $f$), the convergence towards $\hat{\theta}$ is slower and the duality gap does not reach the true suboptimality gap as it should ideally. $f = 10$ provides the best performance.

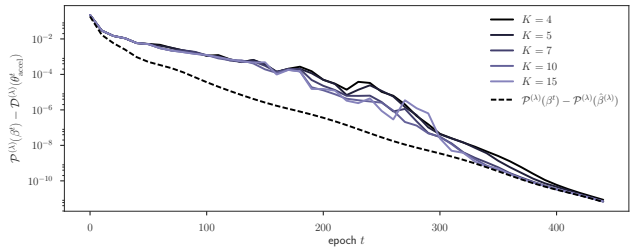

*Figure 7.* Duality gap evaluated with $\theta_{\text{accel}}$ as a function of the parameter $K$, for $f = 10$.

Figure 7 shows that the choice of $K$ is not critical: all performances are nearly equivalent. Hence, we keep the default choice $K = 5$ proposed in Scieur et al. (2016).

### A.2. Working set size policy

In this section, we demonstrate how the growth policy we chose in (14) behaves better than others. We consider two types of growth: geometric of factor $\gamma$:

$$p_t = \min(\gamma |\mathcal{S}_{\beta^{t-1}}|, p) \ , \tag{15}$$

and linear of factor $\gamma$:

$$p_t = \min(\gamma + |\mathcal{S}_{\beta^{t-1}}|, p) \ . \tag{16}$$

We implement these two strategies with factor 2 and 4 for the geometric, and 10 and 50 for the linear. We consider two scenarios:

- *undershooting*, with $p_1 = 10$ much smaller than the true support size $|\mathcal{S}(\hat{\beta})| = 983$ (obtained with $\lambda = \lambda_{\max}/20$),

- *overshooting*, with $p_1 = 500$ much larger than the true support size $|\mathcal{S}(\hat{\beta})| = 63$ (obtained with $\lambda = \lambda_{\max}/5$).

Figure 8 shows that, when the first working set is too small (choice of $p_1 = 10$), the approximate solutions are dense and the subsequent $\mathcal{W}_t$ grow in size. Amongst the four strategies considered, the geometric growth with factor 2 quickly reaches the targeted support size (contrary to the linear strategies), and does not create way too large WS like the geometric strategy with factor 4 does.

Figure 9 shows that, if the initial guess is too large, using $|\mathcal{S}_{\beta^{t-1}}|$ instead of $|\mathcal{W}_{t-1}|$ immediately decreases the size of $\mathcal{W}_1$, thus correcting the initial mistake and avoiding solving too large subproblems.

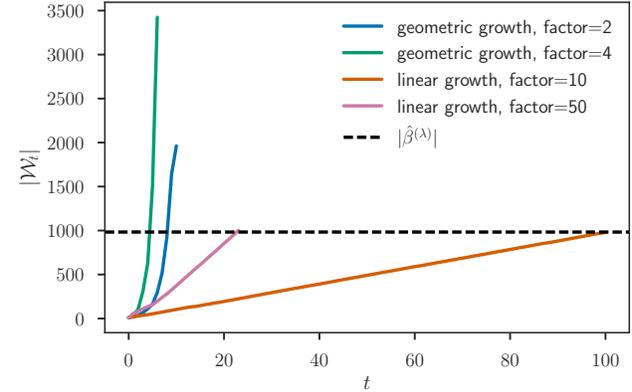

*Figure 8.* Size of working sets defined by CELER with linear or geometric growth, when the support size is underestimated.

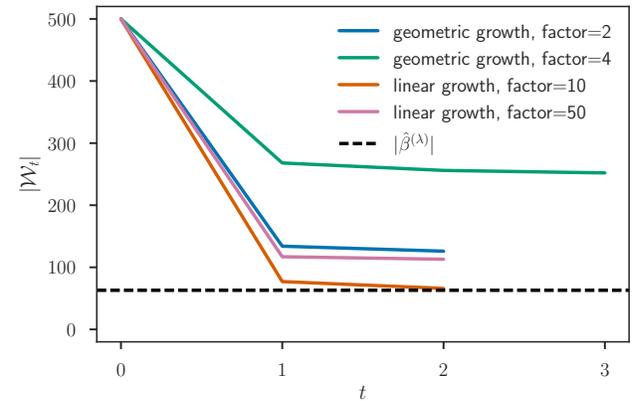

*Figure 9.* Size of working sets defined by CELER with linear or geometric growth, when the support size is overestimated.

### A.3. Path on coarser grid of $\lambda$

We repeat the experiment of Section 6.3 with a grid of 10 values between $\lambda_{\max}$ and $\lambda_{\max}/10^2$. The running time of CELER is still inferior to the one of BLITZ.

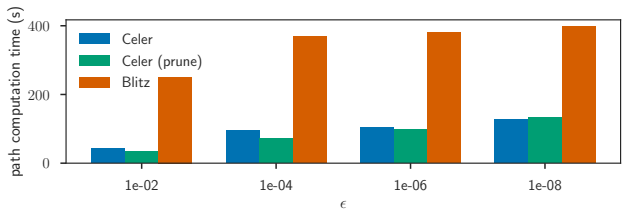

*Figure 10.* Times to solve the Lasso path to precision $\epsilon$ for 10 values of $\lambda$, from $\lambda_{\max}$ to $\lambda_{\max}/100$, on the Finance data. CELER outperforms BLITZ. Both safe and prune versions behave similarly.

### A.4. Path on other dataset

Table 2 reproduces the results of Figure 4 on another dataset: bcTCGA, obtained from the The Cancer Genome Atlas (TCGA) Research Network[12]. For this dense dataset, $n = 536$ and $p = 17,323$ (unregularized intercept column added). The grid goes from $\lambda_{\max}$ to $\lambda_{\max}/100$. The conclusions from Figure 4 still hold.

*Table 2.* Computation time (in seconds) for CELER (no pruning) and BLITZ to reach a given precision $\epsilon$ for a Lasso path on a dense grid, on the bcTCGA dataset.

| $\epsilon$ | $10^{-2}$ | $10^{-4}$ | $10^{-6}$ | $10^{-8}$ |
| --- | --- | --- | --- | --- |
| CELER | 6 | 45 | 160 | 255 |
| BLITZ | 22 | 101 | 252 | 286 |

Note that for the lowest precision the Blitz solver stops running due to internal stopping criterion measuring primal decrease and time spent on working set, but the evaluated duality gap when the solver stops is not always lower than $\epsilon$ along the path.

---

[12] http://cancergenome.nih.gov/